\documentclass[a4paper,UKenglish,cleveref, autoref, thm-restate]{lipics-v2021}

\usepackage{tcolorbox}
\title{Large Multi-modal Model Cartographic Map Comprehension for Textual Locality Georeferencing} 

\titlerunning{LMMs for Textual Locality Georeferencing} 

\author{Kalana Wijegunarathna}{School of Mathematical and Computational Sciences, Massey University, Auckland, New Zealand}{k.wijegunarathna@massey.ac.nz}{https://orcid.org/0000-0001-7458-4801}{}

\author{Kristin Stock}{School of Mathematical and Computational Sciences, Massey University, Auckland, New Zealand}{k.stock@massey.ac.nz}{https://orcid.org/0000-0002-5828-6430}{}

\author{Christopher B. Jones}{School of Computer Science and Informatics, Cardiff University, Cardiff, UK}{jonescb2@cardiff.ac.uk}{https://orcid.org/0000-0001-6847-7575}{}

\authorrunning{K. Wijegunarathna, K. Stock, and C. B. Jones} 

\Copyright{Kalana Wijegunarathna, Kristin Stock, and Christopher B. Jones} 



\funding{This research was partly funded by the Ministry of Business Innovation and Employment Smart Ideas Fund under the BioWhere Project
(grant number MAUX2104).}

\nolinenumbers 

\EventEditors{Katarzyna Sila-Nowicka, Antoni Moore, David O'Sullivan, Benjamin Adams, and Mark Gahegan}
\EventNoEds{5}
\EventLongTitle{13th International Conference on Geographic Information Science (GIScience 2025)}
\EventShortTitle{GIScience 2025}
\EventAcronym{GIScience}
\EventYear{2025}
\EventDate{August 26--29, 2025}
\EventLocation{Christchurch, New Zealand}
\EventLogo{}
\SeriesVolume{346}
\ArticleNo{12}

\begin{document}

\maketitle

\begin{abstract}
Millions of biological sample records collected in the last few centuries archived in natural history collections are un-georeferenced. Georeferencing complex locality descriptions associated with these collection samples is a highly labour-intensive task collection agencies struggle with. None of the existing automated methods exploit maps that are an essential tool for georeferencing complex relations. We present preliminary experiments and results of a novel method that exploits multi-modal capabilities of recent Large Multi-Modal Models (LMM). This method enables the model to visually contextualize spatial relations it reads in the locality description. We use a grid-based approach to adapt these auto-regressive models for this task in a zero-shot setting. Our experiments conducted on a small manually annotated dataset show impressive results for our approach ($\sim$1 km Average distance error) compared to uni-modal georeferencing with Large Language Models and existing georeferencing tools. The paper also discusses the findings of the experiments in light of an LMM's ability to comprehend fine-grained maps. Motivated by these results, a practical framework is proposed to integrate this method into a georeferencing workflow.  
\end{abstract}

\section{Introduction}
\label{sec:intro}

Georeferencing is the process of relating or interpreting information to a geographic location \cite{hill2009georeferencing, chapman2020georeferencing, hackeloeer2014georeferencing}. \textit{Informal} georeferencing is the association of information with a location using place names (also called toponyms) or location descriptions from ordinary human discourse. On the other hand, \textit{formal} georeferencing refers to exact locations using formal quantitative representations such as latitude and longitude coordinates or other spatial referencing systems \cite{hill2009georeferencing}. The task of converting an \textit{informal} georeference to a \textit{formal} georeference can be challenging due to reasons such as colloquial place names, outdated names, historical places, the use of vague relative spatial relations, and differences in place representations in different gazetteers (geospatial databases).


A vast amount of information is locked up in extensive collections of unstructured textual data that is yet to be systematically georeferenced. These collections include but are not limited to web pages, social media articles, academic research articles, biological collection specimen records, and memoirs. The ubiquity of georeferencing has led to numerous georeferencing techniques adopted in various application domains. For example, attempts have been made to georeference social media posts, social media images, satellite and aerial images, web documents, and collection records from natural history archives \cite{wieczorek_point-radius_2004,guo_georeferencing_2008, scherrer2021social, hackeloeer2014georeferencing, melo2017automated}. In this study, we focus on georeferencing textual locality descriptions in records of natural history specimens found in museum and herbarium archives, where it is estimated that of the order of 3 billion records are preserved \cite{arino2010approaches}. It is also estimated that manual georeferencing of  digital records without coordinates held globally could take over 5000 person-years \cite{stock2023biowhere}.

A locality description is a textual description of the location at which a biological or other sample was collected. These descriptions are part of the information recorded about the specimen or sample by the collector and, for millions of pre-GPS collection records, they can be the only detailed information about the collection location.  Georeferencing such locality descriptions for purposes of biodiversity studies is a considerable challenge, especially due to their sheer volume and the descriptions themselves often employing quite complex language with one or more relative spatial relations \cite{marcer2021natural}. Much of the published literature on georeferencing entire sentences has focused on social media posts, with the more advanced methods using various forms of language models including transformer models \cite{vaswani2017attention}. Methods developed for  georeferencing social media posts can also rely heavily on metadata, such as the user network. The locality descriptions with which we are concerned differ significantly from the text of social media postings in their frequent use of relative descriptions often with multiple reference named places, and where the described location is separate (offset) from that of the finer-grained place names. Several studies have focused on the development of methods to georeference such relative locality descriptions in natural history records but little progress has been made to date on the application of current deep learning methods.

\begin{figure}
    \centering
    \includegraphics[width=1\linewidth]{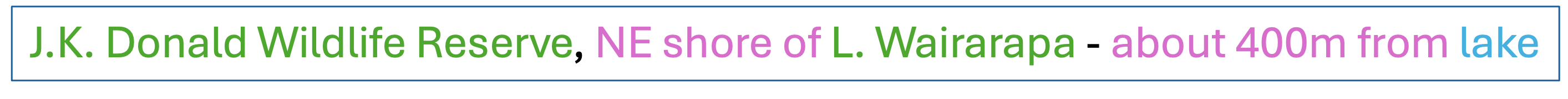}
    \caption{A well defined example locality description from a collection held by the Allen Hebarium\footnotemark. Green and purple indicate place names and relative spatial indicators respectively. Here, "lake" is a coreference to Lake Wairarapa.}
    \label{fig:example}
\end{figure}

Figure \ref{fig:example} provides an example of a locality description. Given this quite specific locality description, a human georeferencer can locate this collection location to a high degree of certainty. Manual georeferencing uses the place names in a locality description to focus on a map that covers the local area to which the description applies. Visualization of the spatial configuration of the named places is vital to a human georeferencer in identifying a point or region of space that appears to correspond to the described location \cite{marcer2021natural}. However, none of the existing automated textual georeferencing methods exploit maps directly. Gazetteer lookup methods only rely primarily on locations of place names, though they can be combined with methods that compute spatial relations \cite{guo_georeferencing_2008, chen_georeferencing_2018}. Current deep learning based methods for georeferencing can use pre-trained language models like BERT \cite{devlin2018bert} that have been pre-trained on masked language modeling and next sentence prediction. They rely on 
fine-tuning these pre-trained models exposing them to large numbers of example texts with their associated locations \cite{scherrer2021social, Li-TransformerPOI-ECIR2023}. Although language models can be adept at learning textual relations, being trained only on language tasks, they do not intrinsically grasp spatial dimensions. The models also do not comprehend spatial extents of the features they are working with. Furthermore, a georeferencing language model trained on one region or country can not be used to infer localities from a different region, requiring more fine-tuning and large volumes of verified data from each region. Additionally, no research appears to have been published to date on using the latest Large Language Models (LLM) for this task.

\footnotetext{https://www.landcareresearch.co.nz/tools-and-resources/collections/allan-herbarium/}

Here we present initial investigations of the potential of Large Multi-Modal models (LMM), that can support tasks combining language and vision, to assist in the georeferencing process for complex locality descriptions. With an LMM's multi-sensory skills, we experiment with a prompting approach that emulates the way that a human might geofererence such descriptions. The intuition in this study is to combine conventional text-based prompting with a map excerpt corresponding to the described location. This exploits the LMM's superior language capabilities while testing its vision encoder for its map reading ability. As current state-of-the-art LMMs excel in language generation and do not perform image segmentation, we superimpose on the map a grid with labelled cells and prompt the LMM to identify the grid cell of the target location. The LMM is given the locality description, the map and the size of the grid cells. We present the results of this study comparing to an existing method, designed for interpreting locality descriptions, and other approaches to using LLMs. Motivated by these results, we design and describe a workflow that can be used to practically automate georeferencing. While the complete workflow is work-in-progress, the core georeferencing module and other elements are already in use for experiments.

Section \ref{sec:related} of the paper will present the related work, after which we will discuss the framework developed to use LMMs in georeferencing in Section \ref{sec:method}. Section \ref{sec:experiments} presents the experiments, results and discussion followed by the conclusion in Section \ref{sec:conclusion}.




\section{Related work}
\label{sec:related}

\subsection{Georeferencing}
The earliest methods for georeferencing text were based on detecting and geocoding place names in the text, which could then be used to assign one or more spatial footprints. Numerous methods for this detection and geocoding process (sometimes referred to jointly as geoparsing) have been developed \cite{gritta_whats_2018, https://wang-Hu-EUPEG-2019}, and some of these have used deep learning approaches. In the case of \cite{gritta-etal-2018-melbourne}, input to a convolutional neural network included the place names, context words and target name, and a vector representation of a pixel map of place name instances, that assisted the disambiguation process.  Document georeferencing methods are currently dominated by language modelling approaches that treat all terms in a text document as evidence for its location \cite{melo2017automated}. The initial language models used Bayesian modelling to associate words with locations, where the locations could be grid cells \cite{serdyukov2009placing, wing2011simple}, or clusters \cite{VanLaere-2014Georeferencing}, where the latter included snapping the location to the most similar already georeferenced existing document (in their case a social media posting). More recently, transformer language models have been adopted either to infer coordinates with a regression approach \cite{scherrer2021social}  or to classify a location as a geographic region \cite{simanjuntak2022we}, or a point of interest \cite{Li-TransformerPOI-ECIR2023}. 

None of the methods above were specifically intended to deal with relative location descriptions such as commonly occur in archived natural history records. Several studies have presented rule-based approaches to georeferencing natural history specimen locality descriptions that use relative spatial relations to specify an offset relative to a reference place name. Different sorts of offset include simply distance from a named object, distance in a specified cardinal direction, and distance along a path. Typically these methods include some or all of detecting place names and spatial relational phrases, disambiguating and hence geocoding the place name, applying the offset distance, and computing some measure of uncertainty. The point radius method \cite{wieczorek_point-radius_2004} was developed to achieve this, in which offsets were calculated relative to a representative point of a feature while also taking account of its extent. The uncertainty of an inferred point-based georeference was expressed as a radial distance that is a function of the six factors of extent of the locality, distance precision, direction precision, unknown datum, coordinate measurement precision and map scale.

The point radius approach was refined in \cite{guo_georeferencing_2008, liu_positioning_2009}, by defining several types of density based uncertainty fields, that take into account the shape of the reference object and which can be combined for complex descriptions. \cite{van_erp_georeferencing_2015} computed distance and direction offsets, accompanied by the spatial minimality toponym disambiguation method \cite{leidner2003grounding}, and applying a confidence measure based on matching the target record to already geofererenced records of the same survey expedition, and to the nearest location of other archived records that have the same species. 

Georeferencing of descriptions of locations that use spatial relations and which were generated in a human subject experiment was described in \cite{chen_georeferencing_2018}. This is one of the few examples of developing and experimenting with geospatial models of spatial relations in natural language expressions outside of the natural history domain. The approach combined models of the applicability of different sorts of relative spatial relations and required the prior existence of a place graph of the spatial relationships between places mentioned in the texts.

\subsection{LMMs and Geospatial Use Cases}

With the recent rapid development of LLMs such as GPT4 \cite{achiam2023gpt}, Llama \cite{touvron2023llama}, PaLM \cite{chowdhery2023palm}, Flamingo \cite{alayrac2022flamingo}, and DeepSeek's V3 \cite{liu2024deepseek} and R1 models \cite{guo2025deepseek}, adding other modalities, including vision, was seen by many as the next improvement. This led to the development of LMMs such as GPT-4Vision \cite{openai20234v}, Qwen-vl \cite{bai2023qwen}, PALM-E \cite{driess2023palm}, Gemini-Pro Vision\footnote{https://aistudio.google.com/}, Sphinx and Janus Pro \cite{chen2025janus}. However, there exist Vision-Language models that predate these LMMs such as CLIP \cite{radford2021learning}, LLaVa \cite{liu2024visual} and BLIP \cite{li2022blip} that combine the two modalities. These models have set benchmarks in various Vision-language tasks such as Visual Question Answering (VQA) \cite{antol2015vqa, kim2024tablevqa}, image captioning \cite{schneider2024m5, nguyen2023improving}, visual language navigation \cite{song2024towards} and visual reasoning \cite{ying2024mmt}. 

LMMs have been applied in several geospatial applications. 
Vision capable models like GPT-4 Vision, Gemini Pro Vision, and Sphinx have been tested for tasks like map element recognition, where GPT4Vision has proved superior \cite{xu2024map}. This study also tests GPT4Vision's comprehension of thematic maps, point pattern, and time series analyses. GPT4Vision has also been tested in its ability to understand weather charts and make forecasts \cite{lawson2025pixels}. Although not using vision capabilities, LLM's abilities to carry out spatial tasks like mapping using code and external tools like MapBox\footnote{https://www.mapbox.com/}, spatial reasoning, and describing interior locations have been tested \cite{ hochmair2024correctness, lipka2024use}. Perhaps the study closest to ours in use case is \cite{zhou2024geolocation}, although they do not use Language-Vision models. This study focuses on geolocating images. They consider maps and image embeddings as two modalities in their multi-modal fusion approach, where they use maps to build a point-cloud representation that can be fused with embeddings from images to exploit heights of buildings to better geolocate images. To the best of our knowledge no method attempts to goereference textual locality descriptions or any form of text documents with LMMs using maps as inputs. We were also  unable to find any literature attempting to georeference textual documents using LLMs.

\section{Methodology}
\label{sec:method}

\begin{figure}
    \centering
    \includegraphics[width=1\linewidth]{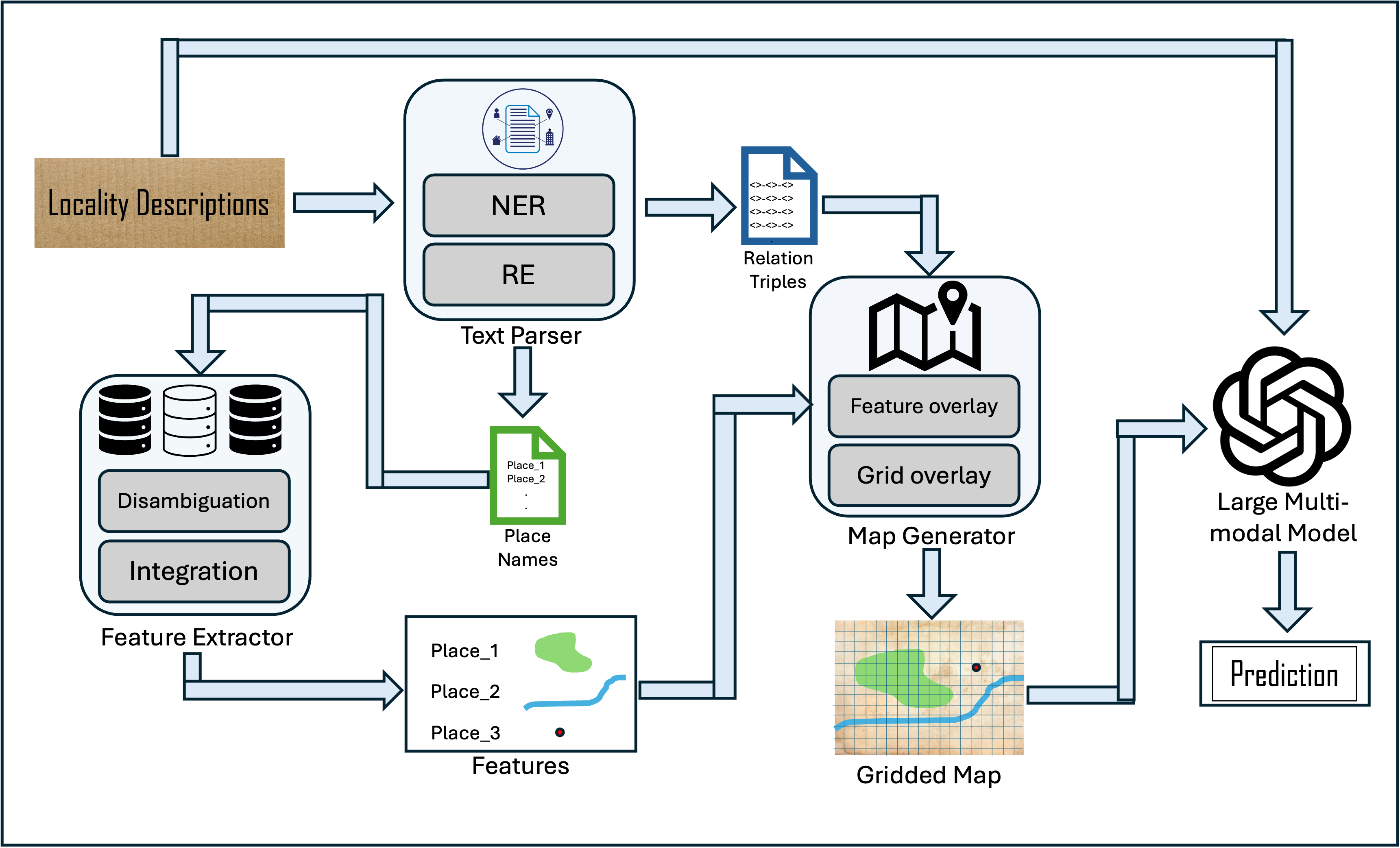}
    \caption{Workflow of the complete automated georeferencing process.}
    \label{fig:workflow}
\end{figure}

Figure \ref{fig:workflow} presents the overall workflow of our proposed framework to utilize large multi-modal models to accurately georeference locality descriptions using gridded maps. We present a detailed description of the proposed method and the individual modules in this section.

\subsection{Textual Information Parsing}

As illustrated in Figure \ref{fig:workflow}, the first step of the process is to extract the names of the places. Grounding named places is the most effective and simplest form of georeferencing and this is vital to our workflow. Named Entity Recognition (NER) \cite{nadeau2009survey} is an extensively researched problem in Natural Language Processing (NLP).  Place names or locations are one of the classical semantic types that NER uses to assign labels to tokens or words \cite{li2020survey}, making most NER solutions accessible for this step of our framework. Off the shelf NER tools such as spaCy\footnote{https://spacy.io/}, StanfordNER \cite{finkel2005incorporating}, NLTK \cite{bird2009natural}, and attention \cite{vaswani2017attention} based pre-trained transformer models \cite{zhong2021frustratingly, su2022global} or modern LLM based approaches \cite{hu2024improving, feng2024note, zhang2024context} can be leveraged for the recognition of place names. Coreference resolution \cite{sukthanker2020anaphora} can be beneficial when parsing relations as illustrated in Figure \ref{fig:example}. The extracted entities are used for Relation Extraction (RE) and finally passed to the Feature Extractor module.

The subsequent step is the extraction of spatial relations between entities. As illustrated in Figure \ref{fig:example}, a single locality description may contain multiple relation clauses in the form of  < \textit{locatum}, \textit{spatial indicator}, \textit{relatum} > triples that relate a location or located object (the \textit{locatum}) to a reference object or location (the \textit{relatum}) with a phrase or clause denoting the spatial relationship (\textit{spatial indicator}). It is also common in locality descriptions to have \textit{degenerate} spatial relations where the locatum is not explicitly mentioned in text but is often the final location being described \cite{Khan-2013}. RE is also a thoroughly studied area. In addition to generic RE methods \cite{zhong2021frustratingly, wan2023gpt, zhang2017review} used in information extraction and NLP, more geospatial relation oriented RE methods have also been developed \cite{kordjamshidi2011spatial, ludwig2016deep}. Relation triples extracted using these methods will then be passed to the Map Generator module (see Figure \ref{fig:workflow}).

\subsection{Geospatial Feature Extractor}

Gazetteers and geospatial databases serve as fundamental resources for the grounding of place names, providing structured and authoritative spatial references. This module will be responsible for extracting relevant features from these knowledge bases, disambiguating them, and selecting the preferred representation of the place instance. While individual states often maintain authoritative gazetteers, several prominent sources provide global coverage. These include, but are not limited to OpenStreetMap\footnote{https://www.openstreetmap.org/} (OSM), GeoNames\footnote{https://www.geonames.org/} and, Google Places API \footnote{https://developers.google.com/maps/documentation/places/web-service}. These sources can vary in their coverage of different place categories (e.g., natural features vs. artificial structures) and in the type of geometric representations they offer, ranging from point-based locations to more complex polygonal and linear footprints. The reliability and completeness of these sources can also vary as some of them are authoritative while others are community-based volunteered information. As the collection country and region are usually included in the records held by collection agencies, we are also able to exploit country-specific gazetteers, allowing us to draw from more authoritative and accurate sources. Conflating these sources provides the most comprehensive set of features for place names mentioned in a locality description. 

First, we query the spatial databases with the place names returned by the previous module. The country name and region of collection can be used for disambiguation. If multiple candidates from the same region from the same source remain, a spatial clustering disambiguation is carried out (\emph{cf} \cite{leidner2003grounding}). This clusters all place names mentioned and selects the candidates that form the strongest cluster, filtering out outliers. Subsequently, we are left with a single feature from each source per place name. In our conflation of sources, we prioritize features with complex geometries as this preserves information like extent and boundaries required for visual georeferencing. Preference is also given to authoritative sources. Finally, the selected features are passed on to the Map Generator module.

\subsection{Map generation}

For the effective application of LMMs in georeferencing, the creation of a map excerpt that is likely to contain the ground truth sample collection location is essential. As the first step of the map generation process, our map server will overlay the features returned from the Feature Extractor on a suitable basemap. Also vital to accurate georeferencing using a vision-based approach is the scale of the map. The map excerpt should be created with all essential landmarks and features necessary for an accurate georeferencing. It should also not be too coarse-grained, to avoid very large grid cells and high uncertainty. We propose the following steps to create the map excerpts:

\begin{enumerate}
    \item In a location description with two or more named places, ${x,y}$ where location $x$ is completely contained in $y$, the full extent of $y$ need not be included in the map extract. Take for example, the following locality description: \textbf{\textit{North Island, Bay of Islands County. Ca 2km north of Puketi.}}
    In this example, North Island contains Bay of Islands County and the county contains Puketi, a small locality. We avoid creating a much coarser grained map by not including the whole extent of the North Island or the Bay of Islands region and focusing on the most fine grained location (Puketi). However, the parent entity is used for disambiguation purposes when retrieving the child entity. 

    \item If there are two or more independent locations at the same level, the map extract must include the full extent of all such features. e.g.: \textbf{\textit{Fiordland, Mount George, south shore of lake at head of Elizabeth Burn, 2km north of peak.}}
    In this example, both Elizabeth Burn and Mount George are included in the map excerpt. The full extent of Fiordland does not need to be included as per 1. above.

    \item If the description includes an absolute distance based spatial relation, we ensure the map excerpt includes a buffered spatial extent of the relatum. 

    \item We ensure features are clearly visible in contrast to the base map. i.e. distinct boundaries for polygon features, clearly highlighted linear and point features. 
    
    \item We ensure legible labels for all identified and retrieved places.  

\end{enumerate}

Subsequently, we superimpose a labeled square grid on the map excerpt. We also record the size of the map grid cells as this is used during inference to calculate relative distances.

\subsection{Multi-modal Georeferencer}

The Georeferencer, essentially a Large Multi-modal Model pre-trained on both language and vision tasks, is the core of the proposed framework. This module takes as input the original locality description that is to be georeferenced along with the gridded map excerpt created by the Map Generator and attempts to predict a grid cell that is most likely to contain the location described in the locality description. Similar to LLMs, LMMs can be sensitive to the prompts used.

\subsubsection{Prompt Design}
\label{sec:promptDesign}

We experimented with several prompts to choose the most effective prompt for this multi-modal georeferencing task. 

\begin{enumerate}
    \item Simple Zero-Shot Prompting \cite{radford2019language}: 
    
    \begin{tcolorbox}[halign=flush left]
        What grid cell/cells represent the following location description?
        
        Location Description:
    \end{tcolorbox}

    \item Automatic Chain-of-thought \cite{zhang2022automatic, wei2022chain}:
    \begin{tcolorbox}[halign=flush left]
        Based on the gridded map given, what grid cell/cells represent the following location description? Think step by step.
        
        Location Description:
    \end{tcolorbox}

    \item Logical Chain-of-Thought Prompting \cite{zhao2023enhancing}: 

    \begin{tcolorbox}[halign=flush left]
        Based on the gridded map given, what grid cell/cells represent the following location description? 
        
        Think step by step. Identify the locations mentioned and use the relative spatial relations mentioned in the description.
        
        Location Description:
    \end{tcolorbox}

    \item Logical Chain-of-Thought Prompting with grid size: 

    \begin{tcolorbox}[halign=flush left]
        Based on the gridded map given, what grid cell/cells represent the following location description? 
        
        Each grid cell is <grid size> $\times$ <grid size>. 
        
        Think step by step. Identify the locations mentioned. If a distance is mentioned in the description, use the grid sizes to calculate the relative distances. 
        
        Location Description:
    \end{tcolorbox}

    \item Persona \cite{white2023prompt} with Logical Chain-of-Thought Prompting with grid size: 

    \begin{tcolorbox}[halign=flush left]
        You are a language and cartography expert.
        Based on the gridded map given, what grid cell/cells represent the following location description? 
        
        Each grid cell is <grid size> $\times$ <grid size>. 
        
        Think step by step. Identify the locations mentioned. If a distance is mentioned in the description, use the grid sizes to calculate the relative distances. 
        
        Location Description:
    \end{tcolorbox}  
    
\end{enumerate}

Our preliminary analysis of these prompting patterns indicated that the Logical Chain-of-thought prompt enhanced with the grid size produced the best results. We will carry out the rest of the experiments with this prompt.

The whole framework proposed in this section is highly reliant on the capability of an LMM to effectively and accurately georeference locations with the aid of a visual map. We present the experiments we conducted to gauge the potential of a multi-modal approach and the merits of diverging from traditional uni-modal text based approaches in the next section.

\section{Experiments}
\label{sec:experiments}
\subsection{Data}

For our preliminary experiment, collection records were obtained from Global
Biodiversity Information Facility\footnote{http://www.gbif.org} (GBIF). GBIF collection records report accurate coordinates for 83\% of the georeferenced records held in it \cite{yesson2007global}. Short location descriptions are more likely to contain only a single place name or a sequence of place names  and no explicit spatial relations (though a comma separated sequence could represent a containment hierarchy). In the absence of descriptive spatial relations, any georeferencing method can, in the best case, only provide the coordinates of the place name mentioned (similar to a gazetteer lookup method). Therefore, the data were first filtered to collect location descriptions that were 60 characters or longer in length, allowing us to gauge the methods' performances on descriptive spatial relations. Given the vast number of collection records and collection types in GBIF, we limited the data to floral specimen collection records from New Zealand provided to GBIF by the Allen Herbarium. The place names and relations were manually annotated as the Text Parser was not implemented at the time of experiment.

For the purposes of this preliminary study, we randomly sampled 25 records to create cartographic map snippets. For this manually curated dataset, we only used OSM to identify named places that are overlaid on the standard OSM base map. For this experiment, we manually checked the excerpts to ensure that the ground truth location was contained within the map excerpt. In our dataset of 25 examples, we observed that the ground truth location was consistently included within the map excerpt generated using the aforementioned steps, without needing any further manual intervention. However, it was observed that in examples with linear features extending over large geographic extents such as highways and rivers, the map excerpt was too coarse grained. In these cases, we manually zoomed in on the non-linear features in the description, making sure to preserve some sections of the linear feature. We will analyse the affects of this manual manipulation in Section \ref{sec:linearFeats}. 

Finally, each data item, $e_i$, in our dataset can be characterised as follows:

\begin{equation}
    e_i = \{text_i, country_i, region_i, map_i, location_i, label_i, scale_i\},
\end{equation}

where $text$ is the locality description, $country$ and $region$ are fields acquired from GBIF, $map$ is the grid-labeled map, $location$ is the ground truth point location of collection as recorded in GBIF  (latitude and longitude coordinate pair), $label$ is the label of the grid cell that contains the $location$ and $scale$ is the size of the grid cell in the map. We manually annotated $label$ for each of these examples after the grid is superimposed. To the best of our knowledge, this is the first publicly available dataset\footnote{https://doi.org/10.6084/m9.figshare.29093882.v1} for fine-grained cartographic map comprehension for LMMs.

\subsection{Baselines}
GeoImp \cite{van2015georeferencing} is perhaps the most recent georeferencing tool for biological specimen georeferencing but unfortunately it is no longer available online. The most effective methods developed for social media post georeferencing (such as Tweets) rely on the metadata and social network information and are therefore unsuitable for our task. GEOLocate \cite{GEOLocate} is an easy-to-use georeferencing system designed specifically for georeferencing natural history collection data, accessible both as a standalone software and an online service. We use this as one of our baselines. GEOLocate enables multiple predictions for each location description, but we only use its best prediction for this study. Additionally, as we are testing the performance of LMMs, we implement our own LLM baselines. All baselines compared against our LMM based generative approach are listed here: 

\begin{enumerate}
    \item \textbf{GEOLocate$_{text}$}: We use GEOLocate's batch processing function over their online service. We only provide the textual description, $text_i$, to the service. 

    \item \textbf{GEOLocate$_{text+region}$}: With this baseline, in addition to the text to georeference, we provide GEOLocate the $country_i$ and $region_i$ from our dataset. 

    \item \textbf{ChatGPT$_{text}$}: Zero-shot georeferencing with OpenAI's ChatGPT\footnote{https://chatgpt.com/}. We use their flagship model, GPT-4o. We manually prompt it adapting a persona prompting pattern \cite{white2023prompt}: 
    \begin{tcolorbox}[halign=flush left]
        You are a language and geography expert. 
        
        Georeference the following location description. Answer with coordinates in decimal degrees. 
        
        Location Description: \{$text_i$\}
    \end{tcolorbox}

    \item \textbf{ChatGPT$_{text+region}$}: This method takes a similar approach to \textbf{ChatGPT}$_{text}$ but enriches the prompt with more context by explicitly providing it with the country and region of collection. 

    \begin{tcolorbox}[halign=flush left]
        You are a language and geography expert. 
        
        Georeference the following location description. Answer with coordinates in decimal degrees. The country and the district of the location are provided. 
        
        This location is in \{$region_i$\}, \{$country_i$\}.
        
        Location Description: \{$text_i$\}
    \end{tcolorbox}

    \item \textbf{GPT-4o$_{text}$}: We use the same prompt as the ChatGPT$_{text}$ and the same underlying model (GPT-4o) but instead of using the web browser, we use the OpenAI's API. The distinction between the two methods is that ChatGPT$_{text}$ has the capability to search the web and retrieve the coordinates of the place names and related information, whereas GPT-4o$_{text}$, accessed via the API, lacks this functionality.

    \item \textbf{GPT-4o$_{text+region}$}: Prompts the GPT-4o model through OpenAI's API using the region and country enhanced prompt as seen in ChatGPT$_{text+region}$.
    
\end{enumerate}

\subsection{Evaluation Metrics}

While distance to ground truth location from the prediction is a straight-forward measure of error for methods that predict coordinates, the measurement of error is slightly more complicated for comparing grid cells with coordinates. We implement three Euclidean distance metrics to calculate the distance error given the correct grid cell label, $label_i$, a predicted grid cell label, $pred_i$, and $scale_i$: 
\begin{equation}
    \begin{array}{ l }
    centroid-distance\ =\sqrt{|x_2 \ -x_1|^{2} +|y_2 -\ y_1|^{2}} \ \times \ scale_{i},
    \end{array}
\end{equation}
\begin{equation}
    \begin{array}{ l }
    max-distance\ =\sqrt{( |x_2 \ -x_1|\ +1)^{2} +( |y_2 -\ y_1|+1)^{2}} \ \times \ scale_{i},
    \end{array}
\end{equation}
\begin{equation}
    min-distance\ =\sqrt{min( ||x_2 \ -x_1|\ -1|,\ |x_2 \ -x_1|)^{2} +min( ||y_2 -\ y_1|-1|,|y_2 -\ y_1|)^{2}} \ \times \ scale_{i},
\end{equation}

where $(x_1, y_1)$ and $(x_2, y_2)$ are two dimensional indices of the grid cells of $label_i$ and $pred_i$, respectively. Each grid cell is a unit square such that $(x_1 y_1), (x_2,y_2) \in \mathbb{N}^{+} \times \mathbb{N}^{+}$. The $centroid-distance$ calculates the Euclidean distance between the two grid cell centroids, where one centroid is considered the ground truth point of collection and the other is the predicted point. The $max-distance$ indicates the upper bound of error, while the $min-distance$ gives the error in the best case scenario. $max-distance$ records an error of $\sqrt{2 \times scale_i^{2}}$ even if both ground truth cell and predicted cell are the same and calculates the distance between the two furthest corners of the given cells. Conversely, $min-distance$ gives an error of zero if the predicted cell and the ground truth cell are the same or are adjacent to each other, calculating the minimum distance between the two cells. For GEOLocate and the generative LLMs, we use the mean Simple Accuracy Error (SAE) between coordinate pairs. We also compare the methods on the percentage of predictions that lie within a 1km, 3km, 10km and $scale_i$ radius of the actual location.

\subsection{Results}

Table \ref{resultsTable} reports the performance of all methods tested. Both methods utilizing \textbf{GEOLocate} produced identical results, signaling that the region and country attributes do not contribute meaningfully to the georeferencing process. This may vary in other regions, such as the United States, where the state-based administrative system is more relevant as indicated in the documentation of GEOLocate. Out of the baselines, \textbf{ChatGPT$_{text+region}$} shows the best results with an average error of 10.12km. \textbf{ChatGPT$_{text}$} follows closely behind with no significant reduction in average distance. This indicates the LLM's ability to disambiguate places to a high degree of accuracy even without the region or country information. \textbf{GPT-4o$_{text}$} produces the highest distance error. However, enhancing the prompt with the region, as in \textbf{GPT-4o$_{text+region}$}, significantly improves results. This suggests the LLM's use of region for disambiguation. The stark difference in performance between the browser versions (ChatGPT$_{text+region}$, ChatGPT$_{text}$) and the same model accessed via the API (GPT-4o$_{text}$, GPT-4o$_{text+region}$) raise an important issue: the inability to browse the web in the API versions significantly hinders the quality of georeferencing. This is also observed in some of the reasoning provided by the model when producing the results. Versions with internet access are able to produce accurate coordinates for named places in the locality descriptions. This also leaves room for further improvement of the LLM based approaches. Providing precise and accurate locations for the named places may result in better quality. However, these improvements are not within scope of this paper. 

Another interesting observation is the change of \% acc at various distances. Although ChatGPT$_{text+region}$ and ChatGPT$_{text}$ produced lower errors (out of the baselines), the \% acc@1km, and \% acc@3km are worse than those of \textbf{GEOLocate} methods. Although able to correctly disambiguate the places and predict within the vicinity, all the LLM based approaches struggle to make a fine-grained prediction. This is to be expected as these methods only predict using point coordinates. Especially for large features such as rivers, mountains, and reserves, a point alone is an inadequate representation for an accurate georeferencing. Furthermore, these results indicate the LLM's inability to take adequate consideration of the rich spatial relations commonly found in these locality descriptions.

The LMM we used for this experiment to test our approach is the OpenAI gpt-4o-2024-08-06 model accessed through their API. As previously discussed in Section \ref{sec:promptDesign}, our prompt for the LMM does not limit the prediction of multiple grid cells. In our experiments, when the model predicts multiple cells, we only consider the first cell predicted. Our proposed approach significantly outperforms the baselines. The centroid-distance of the LMM is an order of magnitude more accurate than the best-performing baseline. Max-distance, which is the upper bound for error given two grid cells, is also markedly lower than all baselines. This indicates our method's ability to consider intricate spatial relations when producing georeferences. When considering a centroid-centroid distance, 60\% of the predictions lie within 1km range of the actual location of collection. This level of accuracy is crucial when manually retrieving biological specimens. 32\% of the predictions made by our multi-modal approach fall exactly in the correct grid cell as the original location. These results clearly demonstrate the significantly superior performance and usefulness of our grid-based multi-modal approach. 

\begin{table}[]
\caption{Average distance errors and percentage of predictions within range of ground truth across the dataset.}
\label{resultsTable}
\begin{tabular}{llccccc}
\hline
\multicolumn{1}{c}{Method}  &          & \begin{tabular}[c]{@{}c@{}}Average \\ distance (km)\end{tabular} & \begin{tabular}[c]{@{}c@{}}\% acc@\\ 1km\end{tabular} & \begin{tabular}[c]{@{}c@{}}\% acc@\\ 3km\end{tabular} & \multicolumn{1}{l}{\begin{tabular}[c]{@{}l@{}}\% acc@\\ 10km\end{tabular}} & \multicolumn{1}{l}{\begin{tabular}[c]{@{}l@{}}\% acc@\\ $scale_i$\end{tabular}} \\ \hline
GEOLocate$_{text}$          &          & 107.23                                                           & 16.0                                                  & 28.0                                                  & 52.0                                                                       & 8.0                                                                             \\
GEOLocate$_{text+region}$   &          & 107.23                                                           & 16.0                                                  & 28.0                                                  & 52.0                                                                       & 8.0                                                                             \\
ChatGPT$_{text}$            &          & 10.91                                                            & 8.0                                                   & 16.0                                                  & 64.0                                                                       & 4.0                                                                             \\
ChatGPT$_{text+region}$     &          & 10.12                                                            & 8.0                                                      & 16.0                                                      & 68.0                                                                           &                                                                                 \\
GPT-4o$_{text}$             &          & 155.82                                                           & 4.0                                                   & 16.0                                                  & 40.0                                                                       & 8.0                                                                             \\
GPT-4o$_{text+region}$      &          & 39.98                                                            & 0                                                     & 12.0                                                  & 56.0                                                                       & 0                                                                               \\ \hline
\multirow{3}{*}{Our method} & min      & 0.42                                                             & 84.0                                                  & 96.0                                                  & 100                                                                        & 88.0                                                                            \\
                            & max      & 2.16                                                             & 24.0                                                  & 80                                                    & 100                                                                        & 0                                                                               \\
                            & centroid & 1.03                                                             & 60.0                                                  & 96.0                                                  & 100                                                                        & 32.0                                                                            \\ \hline
\end{tabular}
\end{table}

\subsection{Discussion}

\subsubsection{Spatial extent and terrain understanding}

\begin{figure}
    \centering
    \includegraphics[width=1\linewidth]{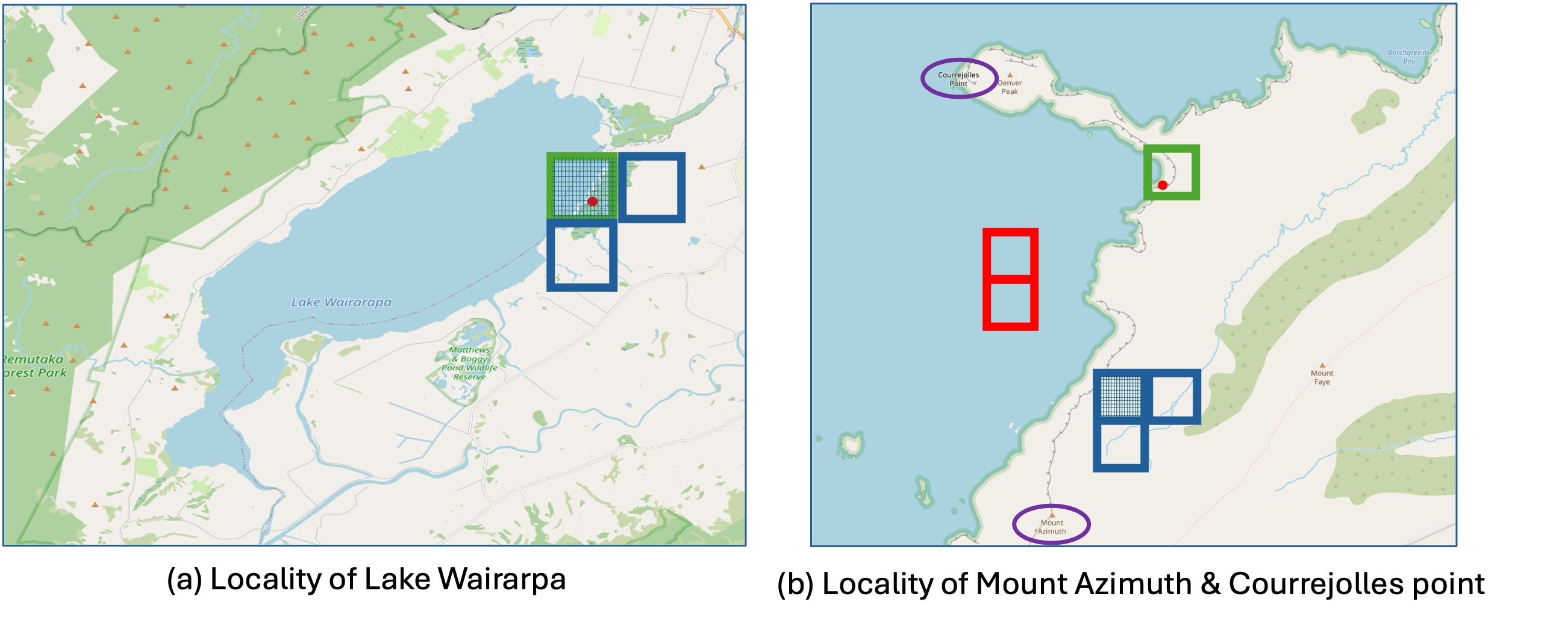}
    \caption{Map excerpts, their labels and their predictions for two locality descriptions: (a) J.K. Donald Wildlife Reserve, NE shore of L. Wairarapa - about 400m from lake \& (b) Mount Azimuth, cliffs between Azimuth and Courrejolles Point near low point in ridge. The grid sizes for (a) and (b) are 1.88km and 0.7 km respectively. The red point indicates the exact point of collection. The green cell indicates the grid cell containing this point. The blue meshed cell indicates the first and primary cell predicted by the model and the other blue cells indicate the secondary predictions. The two place names mentioned in (b) are highlighted for visual clarity and the red cells indicate some of the cells considered during the reasoning of the model.}
    \label{fig:terrain}
\end{figure}

A unique advantage of a multi-modal approach to georeferencing is its potential to understand spatial extents without being limited to simple coordinates. We analyzed the results to identify if the model is indeed capable of understanding extents of features. Figure \ref{fig:terrain}(a) demonstrates an example where the model accurately identified the correct grid cell containing the point of collection. This is the map excerpt and prediction for the locality description shown in Figure \ref{fig:example}. OSM did not find a match for J.K. Donald Wildlife Reserve and the model was restricted to only looking at the lake and its locality. Despite this, the model's ability to correctly predict the grid cell demonstrates the model's ability not only to identify the boundaries of the lake but also the distance from the border where the collection may have taken place (i.e. the "shore" in the locality description). Also of interest is the reasoning it produced for the prediction. The LLM response stated that it considers the green area that looks like a "vegetation patch" to be the J.K. Donald wildlife reserve. This shows the model's ability to identify and reason with topographic features on the base map. Although the LLM's mentioned feature identity is questionable (as OSM's name for that patch is Wairarapa Moana Wetland), this highlights a capability that could be highly beneficial for map-based spatial reasoning. 

Figure \ref{fig:terrain}(b) provides another similar example. In this case, the prediction is far from the actual collection location. However this is understandable when we analyse the locality description: \textbf{\textit{Mount Azimuth, cliffs between Azimuth and Courrejolles Point near low point in ridge}}. Without contour lines or other altitude information, the phrase "low point in ridge" is indiscernible. What is of interest is the calculation the model made for "between". The initial reasoning calculations made by the model predicted the cells marked in red as the cells that represent "between Azimuth and Courrejolles Point". However, it later disregarded these cells in favour of the grids marked in blue. Although not explicitly stated, it seems to have avoided predicting a place in the ocean. This may also have been helped by the mention of an unnamed cliff. This ability of understanding terrain as shown in both examples opens the door to incorporating species-related habitat information into our approach. This could include characteristics such as whether a species inhabits land or water and even probabilistic heat maps on a species' preferential ecosystem.

\subsubsection{Linear Features}
\label{sec:linearFeats}

\begin{figure}
    \centering
    \includegraphics[width=1\linewidth]{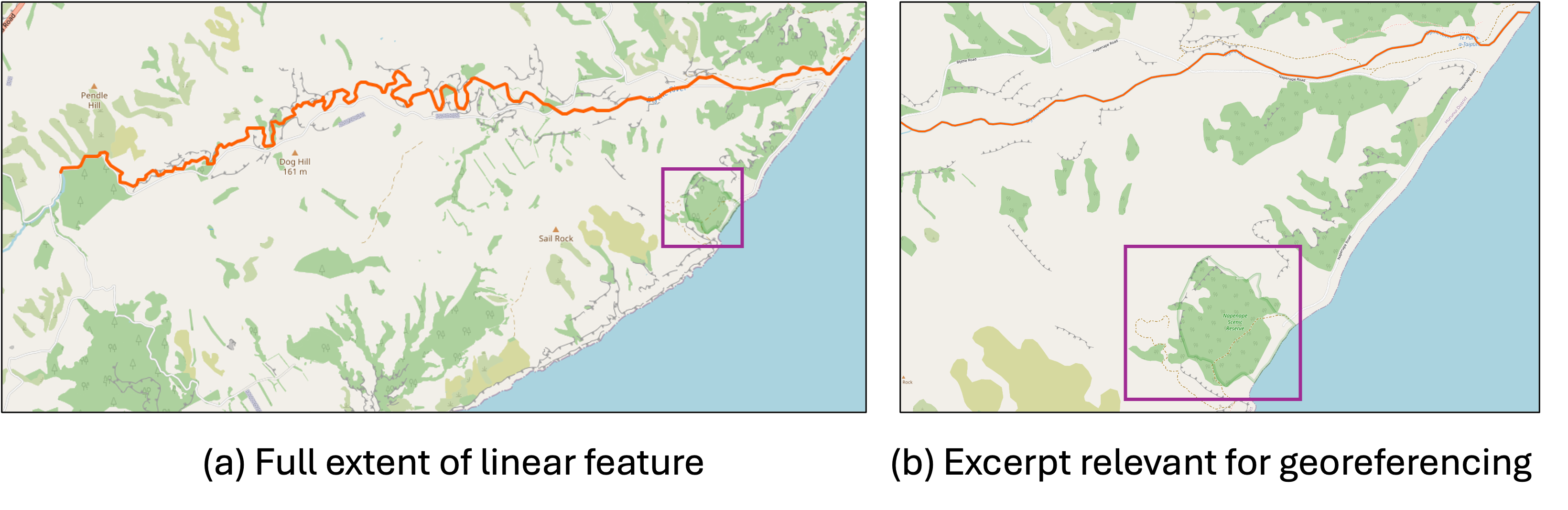}
    \caption{Two map excerpts for the same locality description. The inclusion of the full extent of the river (highlighted in red), as shown in (a) produces a much coarser map compared to (b). The Napenape Scenic Reserve is segmented in purple for visual clarity.}
    \label{fig:linear}
\end{figure}

As mentioned earlier during the creation of the gridded map dataset, manual intervention was needed in the case of linear features. 9 out of the 25 samples contained linear features. Figure \ref{fig:linear} demonstrates this issue, presenting two map excerpts for the following locality description: "\textbf{\textit{North Canterbury, Napenape Scenic Reserve, 3km south of mouth of Blythe River on coast.}}". Including the complete linear feature resulted in a vastly coarser grained map where the subsequently applied grid cells were 1.25km in scale. The map excerpt relevant for the accurate georeferencing would produce much finer grained cells of size 450m, allowing the model to not only pay attention to the river and the reserve but also differentiate grid cells based on whether they lie close to the coast or not. The proposed framework will benefit from further experiments on limiting the extent of the map especially with regard to linear features. A potential avenue is the exploration of distances to the other mentioned features and using these relations to limit the scope of the map. 

Another observation on linear features was the vision encoder's difficulty in comprehending the continuity of the linear features. Some confusion was observed when one road meets another at a junction but continues to be the same road after it. However, this can be remedied by custom labels placed at regular intervals of the linear feature. 

\subsubsection{Enhancing vision models' map comprehension}

Along with the confusion with linear features, we also noticed a tendency of the model to misrecognize the location of a feature using the label on the map instead of the icon or marker. This is contrary to findings in coarser grained maps \cite{xu2024map}. These issues persist due to models like GPT-4o(Vision) not being specifically trained for map comprehension. Despite these inaccuracies, the performance of this zero-shot multi-modal approach is vastly superior to text only approaches. However, there is still space for improvement through fine-tuning, which would the take into account the considerable variation in the forms of locality descriptions. The large numbers of natural history records collected from many different countries around the globe with detailed locality descriptions present an invaluable source of information to fine-tune (or perhaps even use during pre-training) vision models on map comprehension. Maps created using our framework can easily be annotated using existing vision models: thus the framework could be used to create a version of the map with the point of collection prominently marked. Existing multi-modal models can then be used for the labelling ("Which grid cell contains the <Red Marker>?") of these maps. These labels can subsequently be used for fine-tuning vision capabilities of other LMMs using the version of the map where the point of collection is removed. Alternatively, this can be used to pre-train open source vision encoders jointly with smaller open weight LLMs\footnote{Where the weights (parameters) of the LLM model are accessible} to build LMMs specialized in map reading. This framework, of distantly supervised learning with cheap machine annotated data, can be regarded as analogous to masked language modeling or next sequence prediction for uni-modal language models.

\section{Conclusion}
\label{sec:conclusion}

This paper presents a novel method for georeferencing textual locality descriptions using LMMs to combine text understanding with map reading. The accuracy of this method is tested against existing tools and the current state-of-the-art LLMs where our method demonstrates greatly superior results. The distance error improves by an order of magnitude compared to the best baseline. Motivated by these results, a framework and workflow were designed to practically integrate LMMs for the task of georeferencing locality descriptions. Along with the model's unique abilities and current shortcomings, the study also revealed avenues for future research that can be used to build powerful models capable of true map comprehension, taking one more step towards GeoAI. 


\supplement{Multi-modal cartographic georeferencing dataset with labels showing ground truth grid cells for collection location of biological specimens. The dataset also indicates the grid cell size and the textual description provided by the collector and location of collection in decimal latitude and longitudes.}

\supplementdetails[subcategory={}, cite={}, swhid={}]{Dataset}{https://doi.org/10.6084/m9.figshare.29093882.v1}

\bibliography{lipics-v2021-sample-article}

\end{document}